\documentclass[11pt,a4paper]{article}
\usepackage[hyperref]{naaclhlt2018}
\usepackage{times}
\usepackage{latexsym}
\usepackage{graphicx}
\usepackage{stfloats}
\usepackage{url}

\aclfinalcopy 

\title{Modeling Global Syntactic Variation in English \\
	Using Dialect Classification}

\author{Jonathan Dunn \\
  University of Canterbury \\
  Department of Linguistics \\
  {\tt jonathan.dunn@canterbury.ac.nz} }

\date{}

\begin{document}
\maketitle
\begin{abstract}
This paper evaluates global-scale dialect identification for 14 national varieties of English as a means for studying syntactic variation. The paper makes three main contributions: (i) introducing data-driven language mapping as a method for selecting the inventory of national varieties to include in the task; (ii) producing a large and dynamic set of syntactic features using grammar induction rather than focusing on a few hand-selected features such as function words; and (iii) comparing models across both web corpora and social media corpora in order to measure the robustness of syntactic variation across registers.
 \end{abstract}

\section{Syntactic Variation Around the World}

This paper combines grammar induction (Dunn, 2018a, 2018b, 2019) and text classification (Joachims, 1998) to model syntactic variation across national varieties of English. This classification-based approach is situated within the task of dialect identification (Section 2) and evaluated against other baselines for the task (Sections 7 and 8). But the focus is modelling syntactic variation on a global-scale using corpus data. On the one hand, the problem is to use a model of syntactic preferences to predict an author's dialect membership (Dunn, 2018c). On the other hand, the problem is to take a spatially-generic grammar of English that is itself learned from raw text (c.f., Zeman, et al., 2017; Zeman, et al., 2018) and adapt that grammar using dialect identification as an optimization task: which constructions are more likely to occur in a specific regional variety?

Because we want a complete global-scale model, we first have to ask: how many national varieties of English are there? This question, considered in Sections 3 and 4, is essential for determining the inventory of regional varieties that need to be included in the dialect identification task. This paper uses data-driven language mapping to find out where English is consistently used, given web data and Twitter data, in order to avoid the arbitrary selection of dialect areas. This is important for ensuring that each construction in the grammar receives the best regional weighting.

What syntactic features are needed to represent variation in English? As discussed in Section 6, this paper uses grammar induction on a large background corpus to provide a replicable and dynamic feature space in order to avoid arbitrary limitations (e.g., lists of function words). The other side of this problem is to optimize grammar induction for regional dialects by using an identification task to learn regional weights for each part of the grammar: how much does a single generic grammar of English vary across dialects? To what degree does it represent a single dominant dialect?

Finally, a corpus-based approach to variation is restricted to the specific domains or registers that are present in the corpus. To what degree is such a model of variation limited to a specific register? This paper uses both web-crawled corpora and social media corpora to explore the robustness of dialect models across domains (Section 8). Along these same lines, how robust is a model of syntactic variation to the presence of a few highly predictive features? This paper uses unmasking, a method from authorship verification (Koppel, et al., 2007), to evaluate the stability of dialect models over rounds of feature pruning (Section 9).

\section{Previous Work}

Because of its long history as a colonial language (Kachru, 1990), English is now used around the world by diverse national communities. In spite of the global character of English, dialectology and sociolinguistics continue to focus largely on sub-national dialects of English within so-called \textit{inner-circle} varieties (for example, Labov, et al., 2016; Strelluf, 2016; Schreier, 2016; Clark \& Watson, 2016). This paper joins recent work in taking a global approach by using geo-referenced texts to represent national varieties (e.g., Dunn, 2018c; Tamaredo, 2018; Calle-Martin \& Romero-Barranco, 2017; Szmrecsanyi, et al., 2016; Sanders, 2010, 2007; c.f., Davies \& Fuchs, 2015). For example, this study of dialect classification contains \textit{inner-circle} (Australia, Canada, United Kingdom, Ireland, New Zealand, United States), \textit{outer-circle} (India, Malaysia, Nigeria, Philippines, Pakistan, South Africa), and \textit{expanding-circle} (Switzerland, Portugual) varieties together in a single model.

The problem is that these more recent approaches, while they consider more varieties of English, have arbitrarily limited the scope of variation by focusing on a relatively small number of features (Grafmiller \& Szmrecsanyi, 2018; Kruger \& van Rooy, 2018; Schilk \& Schaub, 2016; Collins, 2012). In practical terms, such work uses a smaller range of syntactic representations than comparable work in authorship analysis (c.f., Grieve, 2007; Hirst \& Feiguina, 2007; Argamon \& Koppel, 2013).

From a different perspective, we could view the modelling of dialectal variation as a classification task with the goal of predicting which dialect a sample belongs to. Previous work has draw on many representations that either directly or indirectly capture syntactic patterns (Gamallo, et al., 2016; Barbaresi, 2018; Kreutz \& Daelemans, 2018; Kroon, et al., 2018). Given a search for the highest-performing approach, other work has shown that methods and features without a direct linguistic explanation can still achieve impressive accuracies (McNamee, 2016; Ionescu \& Popescu, 2016; Belinkov \& Glass, 2016; Ali, 2018).

On the other hand, there is a conceptual clash between potentially topic-based methods for dialect identification and other tasks that explicitly model place-specific language use. For example, text-based geo-location can use place-based topics to identify where a document is from (c.f., Wing \& Baldridge, 2014; Hulden, et al., 2015; Lourentzou, et al., 2017). And, at the same time, place-based topics can be used for both characterizing the functions of a location (c.f., Adams \& McKenzie, 2018; Adams, 2015) and disambiguating gazeteers (c.f., Ju, et al., 2016). This raises an important conceptual problem: when does predictive accuracy reflect \textit{dialects} as opposed to either place-references or place-based content? While geo-referenced corpora capture both types of information, syntactic representations focus specifically on \textit{linguistic} variation while place-references and place-based topics are part of document content rather than linguistic structure.

\section{Where Is English Used?}

\begin{table}
	\centering
	\begin{tabular}{|l|r|r|}
		\hline
		\textbf{Region} & \textbf{CC} & \textbf{TW} \\
		\hline
		Africa, North & 123,859,000 & 85,552,000 \\
		Africa, Southern & 59,075,000 & 87,348,000 \\
		Africa, Sub & 424,753,000 & 254,200,000 \\
		\hline
		America, Brazil & 218,119,000 & 118,138,000 \\
		America, Cen. & 886,610,000 & 383,812,000 \\
		America, North & 236,590,000 & 350,125,000 \\
		America, South & 1,163,008,000 & 402,150,000 \\
		\hline
		Asia, Cen. & 965,090,000 & 102,794,000 \\
		Asia, East & 2,201,863,000 & 95,704,000 \\
		Asia, South & 448,237,000 & 331,192,000 \\
		Asia, Southeast & 2,011,067,000 & 245,181,000 \\
		\hline
		Europe, East & 4,553,101,000 & 322,460,000 \\
		Europe, Russia & 101,444,000 & 105,045,000 \\
		Europe, West & 2,422,855,000 & 823,807,000 \\
		\hline
		Middle East & 660,732,000 & 222,985,000 \\
		\hline
		Oceania & 164,025,000 & 213,064,000 \\
		\hline
		\textsc{\textbf{total}} & \textbf{16.65 billion} & \textbf{4.14 billion} \\
		\hline
	\end{tabular}
	\caption{Background Corpus Size in Words by Region}
	\label{tab:1}
\end{table}

The goal of this paper is to model syntactic variation across all major or robust varieties of English. But how do we know which varieties should be included? Rather than select some set of varieties based on convenience, we take a data-driven approach by collecting global web-crawled data and social media data to determine where English is used. This approach is biased towards developed countries with access to digital technologies. As shown in Table 1, however, enough global language data is available from both sources to determine where national varieties of English exist.

Data comes from two sources of digital texts: web pages from the Common Crawl\footnote{\url{http://commoncrawl.org}} and social media from Twitter.\footnote{\url{http://twitter.com}} Both types of data have been used previously to study dialectal and spatial variation in language. More commonly, geo-referenced Twitter data has been taken to represent language-use in specific places (e.g., Eisenstein, et al., 2010; Roller, et al., 2012; Kondor, et al., 2013; Mocanu, et al., 2013; Eisenstein, et al., 2014; Graham, et al., 2014; Donoso \& Sanchez, 2017); regional variation in Twitter usage was also the subject of a shared task at PAN-17 (Rangel, et al., 2017). Web-crawled data has also been curated and prepared for the purpose of studying spatial variation (Goldhahn, et al., 2012; Davies \& Fuchs, 2015), including the use of country-level domains for geo-referencing (Cook \& Brinton, 2017). This paper builds on such previous work by systematically collecting geo-referenced data from both sources on a global scale. The full web corpus is available for download.\footnote{\url{https://labbcat.canterbury.ac.nz/download/?jonathandunn/CGLU_v3}}

For the Common Crawl data (abbreviated as CC), language samples are geo-located using country-specific top-level domains. The assumption is that a language sample from a web-site under the \textit{.ca} domain originated from Canada (c.f., Cook \& Brinton, 2017). This approach to regionalization does not assume that whoever produced that language sample was born in Canada or represents a traditional Canadian dialect group; rather, the assumption is only that the sample represents someone in Canada who is producing language data. Some countries are not available because their top-level domains are used for other purposes (i.e., \textit{.ai}, \textit{.fm}, \textit{.io}, \textit{.ly}, \textit{.ag}, \textit{.tv}). Domains that do not contain geographic information are also removed from consideration (e.g., \textit{.com} sites). The Common Crawl dataset covers 2014 through the end of 2017, totalling 81.5 billion web pages. As shown in Table 1, after processing this produces a corpus of 16.65 billion words.

The basic procedure for processing the Common Crawl data is to look at text within paragraph tags: any document with at least 40 words within paragraph tags from a country-level domain is processed. Noise like navigational items, boilerplate text, and error messages is removed using heuristic searches and also using deduplication: any text that occurs multiple times on the same site or multiple times within the same month is removed. A second round of deduplication is used over the entire dataset to remove texts in the same language that occur in the same country. Its limited scope makes this final deduplication stage possible. For reproducibility, the code used for collecting and processing the Common Crawl data is also made available.\footnote{\url{https://github.com/jonathandunn/common_crawl_corpus}}

The use of country-level domains for geo-referencing raises two questions: First, are there many domains that are not available because they are not used or are used for non-geographic purposes? After removing irrelevant domains like \textit{.tv}, the CC dataset covers 166 countries (30 of which are not included in the Twitter corpus) while the Twitter corpus covers 169 countries (33 of which are not included in the CC corpus). Thus, while the use of domains does remove some countries from consideration, the effect is limited. Second, does the amount of data for each country domain reflect the actual number of web pages from that country? In other words, some countries like the United States are less likely to use their top-level codes. However, the United States is still well-represented in the model. The bigger worry is that regional varieties from Africa or East Asia, both of which are under-represented in these datasets, might be missing from the model.

For the Twitter corpus, a spatial search is used to collect Tweets from within a 50km radius of 10k cities.\footnote{\url{https://github.com/datasets/world-cities}} Such a search avoids biasing the selection by using language-specific keywords or hashtags. The Twitter data covers the period from May of 2017 until early 2019. This creates a corpus containing 1,066,038,000 Tweets. The language identification component, however, only provides reliable predictions for samples containing at least 50 characters. Thus, the corpus is pruned to include only those Tweets above that length threshold. As shown in Table 1, this produces a corpus containing 4.14 billion words with a global distribution. Language identification (LID) is important here because a failure to identify some regional varieties of English will ultimately bias the model. The LID system used is available for testing.\footnote{\url{https://github.com/jonathandunn/idNet/}} But given that the focus is a major language, English, the performance of LID is not a significant factor in the overall model of syntactic variation.

The datasets summarized in Table 1 include many languages other than English. The purpose is to provide background information about where robust varieties of English are found: where is English discovered when the search is not biased by looking only for English? On the one hand, some regions may be under-represented in these datasets; if national varieties are missing from a region, it could be (i) that there is no national variety of English or (ii) that there is not enough data available from that region. On the other hand, Table 1 shows that each region is relatively well-represented, providing confidence that we are not missing other important varieties.

\section{How Many Varieties of English?}

We take a simple threshold-based approach to the question of which regional varieties to include: any national variety that has at least 15 million words in both the Common Crawl and Twitter datasets is included in the attempt to model all global varieties of English. This threshold is chosen in order to ensure that sufficient training/testing/development samples are available for each variety. The inventory of national varieties in Table 2 is entirely data-driven and does not depend on distinctions like dialects vs. varieties, inner-circle vs. outer-circle, or native vs. non-native. Instead, the selection is empirical: any area with a large amount of observed English usage is assumed to represent a regional variety. Since the regions here are based on national boundaries, we call these national varieties. We could just as easily call them national dialects. 

\begin{table}
\centering
\begin{tabular}{|l|r|r|}
\hline
\textbf{Country} & \textbf{CC} & \textbf{TW} \\
\hline
South Africa & 53,447,000 & 57,017,000 \\
Nigeria & 113,957,000 & 29,390,000 \\
\hline
Canada & 149,882,000 & 97,835,000 \\
United States & 42,890,000 & 220,947,000 \\
\hline
India & 71,219,000 & 80,038,000 \\
Pakistan & 140,190,000 & 34,044,000 \\
\hline
Malaysia & 198,566,000 & 18,296,000 \\
Philippines & 209,476,000 & 19,705,000 \\
\hline
England & 62,811,000 & 43,376,000 \\
Ireland & 43,975,000 & 46,045,000 \\
\hline
Portugual & 20,960,000 & 23,333,000 \\
Switzerland & 15,459,000 & 17,788,000 \\
\hline
Australia & 29,129,000 & 98,955,000 \\
New Zealand & 87,951,000 & 37,428,000 \\
\hline
\textsc{\textbf{total}} & \textbf{1.23 billion} & \textbf{0.82 billion} \\
\hline
  \end{tabular}
  \caption{English Varieties by Dataset in N. Words}
  \label{tab:1}
\end{table}

Nevertheless, the inventory (sorted by region) contains within it some important combinations. There are two African varieties, two south Asian varieties, two southeast Asian varieties, two native-speaker European varieties and two non-native-speaker European varieties. Taken together, these pairings provide a rich ground for experimentation. Are geographically closer varieties more linguistically similar? Is there an empirical reality to the distinction between inner-circle and outer-circle varieties (e.g., American English vs. Malaysian English)? The importance of this language-mapping approach is that it does not assume the inventory of regions.

\section{Data Preparation and Division}

The goal of this paper is to model syntactic variation using geo-referenced documents taken from web-crawled and social media corpora. Such geo-referenced documents represent language use \textit{in} a particular place but, unlike traditional dialect surveys, there is no assurance that individual authors are native speakers \textit{from} that place. We have to assume that most language samples from a given country represent the native English variety of that country. For example, many non-local residents live in Australia; we only have to assume that \textit{most} speakers observed in Australia are locals. 

In order to average out the influence of out-of-place samples, we use random aggregation to create samples of exactly 1,000 words in both corpora. For example, in the Twitter corpus this means that an average of 59 individual Tweets from a place are combined into a single sample. First, this has the effect of providing more constructions per sample, making the modeling task more approachable. Second and more importantly, individual out-of-place Tweets are reduced in importance because they are aggregated with other Tweets presumably produced by local speakers.

\begin{table}
\centering
\begin{tabular}{|l|r|r|}
\hline
\textbf{~} & \textbf{CC} & \textbf{TW} \\
\hline
Training Samples & 327,500 & 308,000 \\
Testing Samples & 66,500 & 64,000 \\
\hline
  \end{tabular}
  \caption{Samples by Function and Dataset}
  \label{tab:1}
\end{table}

The datasets are formed into training, testing, and development sets as follows: First, 2k samples are used for development purposes regardless of the amount of data from a given regional variety. Depending on the size of each variety, at least 12k training and 2.5k testing samples are available. Because some varieties are represented by much larger corpora (i.e., Tweets from American English), a maximum of 25k training samples and 5k testing samples are allowed per variety per register. This creates a corpus with 327,500 training and 66,500 testing samples (CC) and a corpus with 308,000 training and 64,000 testing samples (TW). As summarized in Table 3, these datasets contain significantly more observations than have been used in previous work (c.f., Dunn, 2018c).

\section{Learning the Syntactic Feature Space}

Past approaches to syntactic representation for this kind of task used part-of-speech n-grams (c.f., Hirst \& Feiguina, 2007) or lists of function words (c.f., Argamon \& Koppel, 2013) to indirectly represent grammatical patterns. Recent work (Dunn, 2018c), however, has introduced the use of a full-scale syntactic representations based on grammar induction (Dunn, 2017, 2018a, 2019) within the Construction Grammar paradigm (CxG: Langacker, 2008; Goldberg, 2006). The idea is that this provides a replicable syntactic representation.

A CxG, in particular, is useful for text classification tasks because it is organized around complex \textit{constructions} that can be quantified using frequency. For example, the ditransitive construction in (1) is represented using a sequence of slot-constraints. Some of these slots have syntactic fillers (i.e., \textsc{noun}) and some have joint syntactic-semantic fillers (i.e., \textit{V:transfer}). Any utterance, as in (2) or (3), that satisfies these slot-constraints counts as an example or instance of the construction. This provides a straight-forward quantification of a grammar as a one-hot encoding of construction frequencies.

~

(1) [\textsc{noun} -- \textit{V:transfer} -- \textit{N:animate} -- \textsc{noun}]
~

(2) ``He mailed Mary a letter."
~

(3) ``She gave me a hand."

~

This paper compares two learned CxGs: first, the same grammar used in previous work (Dunn, 2018c); second, a new grammar learned with an added association-based transition extraction algorithm (Dunn, 2019). These are referred to as CxG-1 (the frequency-based grammar in Dunn, 2019) and CxG-2 (the association-based grammar), respectively. Both are learned from web-crawled corpora separate from the corpora used for modeling regional varieties (from Baroni, et al., 2009; Majl\c{i}s \& \v{Z}abokrtsk\'{y}, 2012; Benko, 2014; and the data provided for the CoNLL 2017 Shared Task: Ginter, et al., 2017). The exact datasets used are available.\footnote{\url{https://labbcat.canterbury.ac.nz/download/?jonathandunn/CxG_Data_FixedSize}}

In both cases a large background corpus is used to represent syntactic constructions that are then quantified in samples from regional varieties. The grammar induction algorithm itself operates in folds, optimizing grammars against individual test sets and then aggregating these fold-specific grammars at the end. This creates, in effect, one large umbrella-grammar that potentially over-represents a regional dialect. From the perspective of the grammar, we can think of false positives (the umbrella-grammar contains constructions that a regional dialect does not use) and false negatives (the umbrella-grammar is missing constructions that are important to a regional dialect). For dialect identification as a task, only missing constructions will reduce prediction performance.

\begin{table}
	\centering
	\begin{tabular}{|l|c|c|}
		\hline
		\textbf{Country} & \textbf{CxG-1 (CC)} & \textbf{CxG-2 (CC)} \\
		\hline
		South Africa & +4.42\% & +4.62\% \\
		Nigeria & -0.93\% & -0.78\% \\
		\hline
		Canada & +4.03\% & +5.17\% \\
		United States & -0.98\% & -1.90\% \\
		\hline
		India & -3.15\% & -10.38\% \\
		Pakistan & -4.76\% & -17.25\% \\
		\hline
		Malaysia & -3.39\% & -11.51\% \\
		Philippines & -4.48\% & -17.39\% \\
		\hline
		England & +4.59\% & +13.98\% \\
		Ireland & +4.26\% & +18.62\% \\
		\hline
		Portugual & -5.82\% & -4.70\% \\
		Switzerland & +0.98\% & +13.96\% \\
		\hline
		Australia & +3.75\% & +8.15\% \\
		New Zealand & +1.83\% & -0.59\% \\
		\hline
	\end{tabular}
	\caption{Relative Average Feature Density}
	\label{tab:1}
\end{table}

How well do CxG-1 and CxG-2 represent the corpora from each regional variety? While prediction accuracies are the ultimate evaluation, we can also look at the average frequency across all constructions for each national dialect. Because the samples are fixed in length, we would expect the same frequencies across all dialects. On the other hand, false positive constructions (which are contained in the umbrella-grammar but do not occur frequently in a national dialect) will reduce the overall feature density for that dialect. Because the classification results do not directly evaluate false positive constructions, we investigate this in Table 4 using the average feature density: the total average frequency per sample, representing how many syntactic constructions from the umbrella-grammar are present in each regional dialect. This is adjusted to show differences from the average for each grammar (i.e., CxG-1 and CxG-2 are each calculated independently).

First, CxG-1 has a smaller range of feature densities, with the lowest variety (Portugal English) being only 10.41\% different from the highest variety (UK English). This range is much higher for CxG-2, with a 36.01\% difference between the lowest variety (Philippines English) and the highest variety (Irish English). One potential explanation for the difference is that CxG-2 is a better fit for the inner-circle dominated training data. This is a question for future work. For now, both grammars pattern together in a general sense: the highest feature density is found in UK English and varieties more similar to UK English (Ireland, Australia). The lowest density is found in under-represented varieties such as Portugal English or Philippines English. Any grammar-adaptation based on dialect identification will struggle to add unknown constructions from these varieties.

\section{Modeling National Varieties}

The main set of experiments uses a Linear Support Vector Machine (Joachims, 1998) to classify dialects using CxG features. Parameters are tuned using the development data. Given the general robust performance of SVMs in the literature relative to other similar classifiers on variation tasks (c.f., Dunn, et al., 2016), we forego a systematic evaluation of classifiers.

\begin{table}
	\centering
	\begin{tabular}{|l|c|c|c|}
		\hline
		\textbf{Features} & \textbf{Prec.} & \textbf{Recall} & \textbf{F1} \\
		\hline
		CxG-1 (CC) & 0.80 & 0.80 & 0.80 \\
		CxG-1 (TW) & 0.75 & 0.76 & 0.76 \\
		\hline
		CxG-2 (CC) & 0.96 & 0.96 & 0.96 \\
		CxG-2 (TW) & 0.92 & 0.92 & 0.92 \\
		\hline
		Funct. (CC) & 0.65 & 0.65 & 0.65 \\
		Funct. (TW) & 0.56 & 0.57 & 0.55 \\
		\hline
		Unigrams (CC) & 1.00 & 1.00 & 1.00 \\
		Unigrams (TW) & 1.00 & 1.00 & 1.00 \\
		\hline
		Bigrams (CC) & 0.98 & 0.98 & 0.98 \\
		Bigrams (TW) & 0.97 & 0.97 & 0.97 \\
		\hline
		Trigrams (CC) & 0.87 & 0.87 & 0.87 \\
		Trigrams (TW) & 0.82 & 0.82 & 0.82 \\
		\hline
	\end{tabular}
	\caption{Classification Performance By Feature Set}
	\label{tab:1}
\end{table}

We start, in Table 5, with an evaluation of baselines by feature type and dataset. We have two general types of features: purely syntactic representations (CxG-1, CxG-2, Function words) and potentially topic-based features (unigrams, bigrams, trigrams). The highest performing feature on both datasets is simple lexical unigrams, at 30k dimensions. We use a hashing vectorizer to avoid a region-specific bias: the vectorizer does not need to be trained or initialized against a specific dataset so there is no chance that one of the varieties will be over-represented in determining which n-grams are included. But this has the side-effect of preventing the inspection of individual features. Vectors for all experiments are available, along with the trained models that depend on these vectors.\footnote{\url{https://labbcat.canterbury.ac.nz/download/?jonathandunn/VarDial_19}}

As \textit{n} increases, n-grams tend to represent structural rather than topical information. In this case, performance decreases as \textit{n} increases. We suggest that this decrease provides an indication that the performance of unigrams is based on location-specific content (e.g., ``Chicago" vs. ``Singapore") rather than on purely linguistic lexical variation (e.g., ``jeans" vs. ``denim"). How do we differentiate between predictions based on place-names, those based on place-specific content, and those based on dialectal variation? That is a question for future work. For example, is it possible to identify and remove location-specific content terms? Here we focus instead on using syntactic representations that are not subject to such interference.

Within syntactic features, function words perform the worst on both datasets with F1s of 0.65 and 0.55. This is not surprising because function words in English do not represent syntactic structures directly; they are instead markers of the types of structures being used. CxG-1 comes next with F1s of 0.80 and 0.76, a significant improvement over the function-word baseline but not approaching unigrams. Note that the experiments using this same grammar in previous work (Dunn, 2018c) were applied to samples of 2k words each. Finally, CxG-2 performs the best, with F1s of 0.96 and 0.92, falling behind unigrams but rivaling bigrams and surpassing trigrams. Because of this, the more detailed experiments below focus only on the CxG-2 grammar.

\begin{table*}
	\centering
	\begin{tabular}{|l|ccc|ccc|}
		\hline
		\textbf{Country} & \textbf{Prec. (CC) } & \textbf{Recall (CC)} & \textbf{F1 (CC)} & \textbf{Prec. (TW) } & \textbf{Recall (TW)} & \textbf{F1 (TW)}\\
		\hline
		South Africa & 0.94 & 0.96 & 0.95 & 0.92 & 0.94 & 0.93 \\
		Nigeria & 0.98 & 0.98 & 0.98 & 0.94 & 0.95 & 0.94 \\
		\hline
		Canada & 0.94 & 0.94 & 0.94 & 0.84 & 0.79 & 0.81 \\
		United States & 0.93 & 0.95 & 0.94 & 0.85 & 0.89 & 0.87 \\
		\hline
		India & 0.97 & 0.98 & 0.97 & 0.97 & 0.97 & 0.97 \\
		Pakistan & 1.00 & 0.99 & 0.99 & 0.98 & 0.98 & 0.98 \\
		\hline
		Malaysia & 0.96 & 0.96 & 0.96 & 0.99 & 0.99 & 0.99 \\
		Philippines & 0.98 & 0.97 & 0.98 & 0.98 & 0.98 & 0.98 \\
		\hline
		England & 0.95 & 0.95 & 0.95 & 0.87 & 0.90 & 0.89 \\
		Ireland & 0.97 & 0.97 & 0.97 & 0.95 & 0.95 & 0.95 \\
		\hline
		Portugual & 0.99 & 0.98 & 0.98 & 0.93 & 0.90 & 0.92 \\
		Switzerland & 0.97 & 0.94 & 0.96 & 0.98 & 0.97 & 0.97 \\
		\hline
		Australia & 0.97 & 0.96 & 0.97 & 0.82 & 0.83 & 0.83 \\
		New Zealand & 0.91 & 0.92 & 0.91 & 0.92 & 0.90 & 0.91 \\
		\hline
		\textsc{\textbf{w. avg}} & \textbf{0.96} & \textbf{0.96} & \textbf{0.96} & \textbf{0.92} & \textbf{0.92} & \textbf{0.92} \\
		\hline
	\end{tabular}
	\caption{Within-Domain Classification Performance (CxG-2)}
	\label{tab:1}
\end{table*}

A closer look at both datasets by region for CxG-2 is given in Table 6. The two datasets (web-crawled and social media) present some interesting divergences. For example, Australian English is among the better performing varieties on the CC dataset (F1 = 0.97) but among the worst performing varieties on Twitter (F1 = 0.83). This is the case even though the variety we would assume would be most-often confused with Australian English (New Zealand English) has a stable F1 across domains (both are 0.91). An examination of the confusion matrix (not shown), reveals that errors between New Zealand and Australia are similar between datasets but that the performance of Australian English on Twitter data is reduced by confusion between Australian and Canadian English.

In Table 4 we saw that the umbrella-grammar (here, CxG-2) better represents inner-circle varieties, specifically UK English and more closely related varieties. This is probably an indication of the relative representation of the different varieties used to train the umbrella-grammar: grammar induction will implicitly model the variety it is exposed to. It is interesting, then, that less typical varieties like Pakistan English and Philippines English (which had lower feature densities) have higher F1s in the dialect identification task. On the one hand, the syntactic differences between these varieties and inner-circle varieties means that the umbrella-grammar misses some of their unique constructions. On the other hand, their greater syntactic difference makes these varieties easier to identify: they are more distinct in syntactic terms even though they are less well represented.

Which varieties are the most similar syntactically given this model? One way to quantify similarity is using errors: which varieties are the most frequently confused? American and Canadian English have 221 misclassified samples (CC), while Canadian and UK English are only confused 36 times. This reflects an intuition that Canadian English is much more similar to American English than it is to UK English. New Zealand and Australian English have 101 misclassifications (again, on CC); but New Zealand and South African English have 266. This indicates that New Zealand English is more syntactically similar to South African English than to Australian English. However, more work on dialect similarity is needed to confirm these findings across different datasets.
 
\section{Varieties on the Web and Social Media}

How robust are models of syntactic variation across domains: in other words, does web-crawled data provide the same patterns as social media data? We conduct two types of experiments to evaluate this: First, we take dialect as a cross-domain phenomenon and train/test models on both datasets together, ignoring the difference between registers. Second, we evaluate models trained entirely on web-crawled data against testing data from social media (and vice-versa), evaluating a single model across registers. The point is to evaluate the impact of registers on syntactic variation: does Australian English have the same profile on both the web and on Twitter?

\begin{table*}[t]
	\centering
	\begin{tabular}{|l|ccc|ccc|}
		\hline
		\textbf{Country} & \textbf{Prec. (CC)} & \textbf{Recall (CC)} & \textbf{F1 (CC)} & \textbf{Prec. (TW)} & \textbf{Recall (TW)} & \textbf{F1 (TW)} \\
		\hline
		South Africa & 0.88 & 0.06 & 0.10 & 0.68 & 0.31 & 0.43 \\
		Nigeria & 0.43 & 0.84 & 0.57 & 0.73 & 0.41 & 0.52 \\
		\hline
		Canada & 0.48 & 0.14 & 0.22  & 0.49 & 0.27 & 0.35 \\
		United States & 0.20 & 0.87 & 0.32 & 0.83 & 0.16 & 0.27 \\
		\hline
		India & 0.65 & 0.94 & 0.77 & 0.38 & 0.90 & 0.54 \\
		Pakistan & 0.96 & 0.41 & 0.58 & 0.88 & 0.36 & 0.51 \\
		\hline
		Malaysia & 0.45 & 0.93 & 0.61 & 0.98 & 0.05 & 0.10 \\
		Philippines & 0.73 & 0.61 & 0.66 & 0.87 & 0.22 & 0.35 \\
		\hline
		England & 0.89 & 0.01 & 0.03 & 0.48 & 0.44 & 0.46 \\
		Ireland & 0.94 & 0.21 & 0.35 & 0.78 & 0.52 & 0.62 \\
		\hline
		Portugual & 0.02 & 0.00 & 0.00 & 0.22 & 0.17 & 0.19 \\
		Switzerland & 0.92 & 0.04 & 0.07 & 0.12 & 0.80 & 0.20 \\
		\hline
		Australia & 0.89 & 0.00 & 0.00 & 0.33 & 0.66 & 0.44 \\
		New Zealand & 0.27 & 0.53 & 0.36 & 0.64 & 0.40 & 0.49 \\
		\hline
		\textsc{\textbf{w. avg}} & \textbf{0.62} & \textbf{0.40} & \textbf{0.33} & \textbf{0.62} & \textbf{0.40} & \textbf{0.40}  \\
		\hline
	\end{tabular}
	\caption{Cross-Domain Models, Trained on CC (Left) and Trained on TW (Right), CxG-2}
	\label{tab:1}
\end{table*}

\begin{table}
	\centering
	\begin{tabular}{|l|c|c|c|}
		\hline
		\textbf{Country} & \textbf{Prec.} & \textbf{Recall} & \textbf{F1} \\
		\hline
		South Africa & 0.91 & 0.92 & 0.92 \\
		Nigeria & 0.94 & 0.95 & 0.95 \\
		\hline
		Canada & 0.87 & 0.84 & 0.85 \\
		United States & 0.85 & 0.90 & 0.87 \\
		\hline
		India & 0.96 & 0.97 & 0.97 \\
		Pakistan & 0.98 & 0.98 & 0.98 \\
		\hline
		Malaysia & 0.97 & 0.96 & 0.96 \\
		Philippines & 0.97 & 0.97 & 0.97 \\
		\hline
		England & 0.87 & 0.90 & 0.89 \\
		Ireland & 0.94 & 0.95 & 0.95 \\
		\hline
		Portugual & 0.94 & 0.90 & 0.92 \\
		Switzerland & 0.96 & 0.93 & 0.95 \\
		\hline
		Australia & 0.87 & 0.86 & 0.87 \\
		New Zealand & 0.89 & 0.87 & 0.88 \\
		\hline
		\textsc{\textbf{w. avg}} & \textbf{0.92} & \textbf{0.92} & \textbf{0.92} \\
		\hline
	\end{tabular}	
	\caption{Single-Set Classification Performance}
	\label{tab:1}
\end{table}

Starting with the register-agnostic experiments, Table 8 shows the classification performance if we lump all the samples into a single dataset (however, the same training and testing data division is still maintained). The overall F1 is the same as the Twitter-only results in Table 6. On the other hand, varieties like Australian English that performed poorly in Twitter perform somewhat better under these conditions. Furthermore, the observation made above that outer-circle varieties are more distinct remains true: the highest performing varieties are the least proto-typical (i.e., Indian English and Philippines English).

But a single model does not perform well across the two datasets, as shown in Table 7. The model trained on Twitter data does perform somewhat better than its counterpart, but in both cases there is a significant drop in performance. On the one hand, this is not surprising given differences in the two registers: we expect some reduction in classification performance across domains like this. For example, the unigram baseline suffers a similar reduction to F1s of 0.49 (trained on CC) and 0.55 (trained on Twitter). 

On the other hand, we would have more confidence in this model of syntactic variation if there was a smaller drop in accuracy. How can we better estimate grammars and variations in grammars across these different registers? Is it a problem of sampling different populations or is there a single population that is showing different linguistic behaviours? These are questions for future work.

\section{Unmasking Dialects}

How robust are classification-based dialect models to a small number of highly predictive features? A high predictive accuracy may disguise a reliance on just a few syntactic variants. Within authorship verification, unmasking has been used as a meta-classification technique to measure the depth of the difference between two text types (Koppel, et al., 2007). The technique uses a linear classifier to distinguish between two texts using chunks of the texts as samples. Here we distinguish between dialects with individual samples as chunks. After each round of classification, the most predictive features are removed. In this case, the highest positive and negative features for each regional dialect are removed for the next classification round. Figure 1 shows the unmasking curve over 100 rounds using the F1 score. Given that there are 14 regional dialects in the model, Figure 1 represents the removal of approximately 2,800 features.

For both datasets, the unigram baseline degrades less quickly than the syntactic model. On the one hand, it has significantly more features in total, so that there are more features to support the classification. On the other hand, given that the most predictive features are being removed, this shows that the lexical model has a deeper range of differences available to support classification than the syntactic model. Within the syntactic models, the classifier trained on web-crawled data degrades less quickly than the Twitter model and maintains a higher performance throughout. 

\begin{figure*}[t]
\includegraphics[scale=0.95]{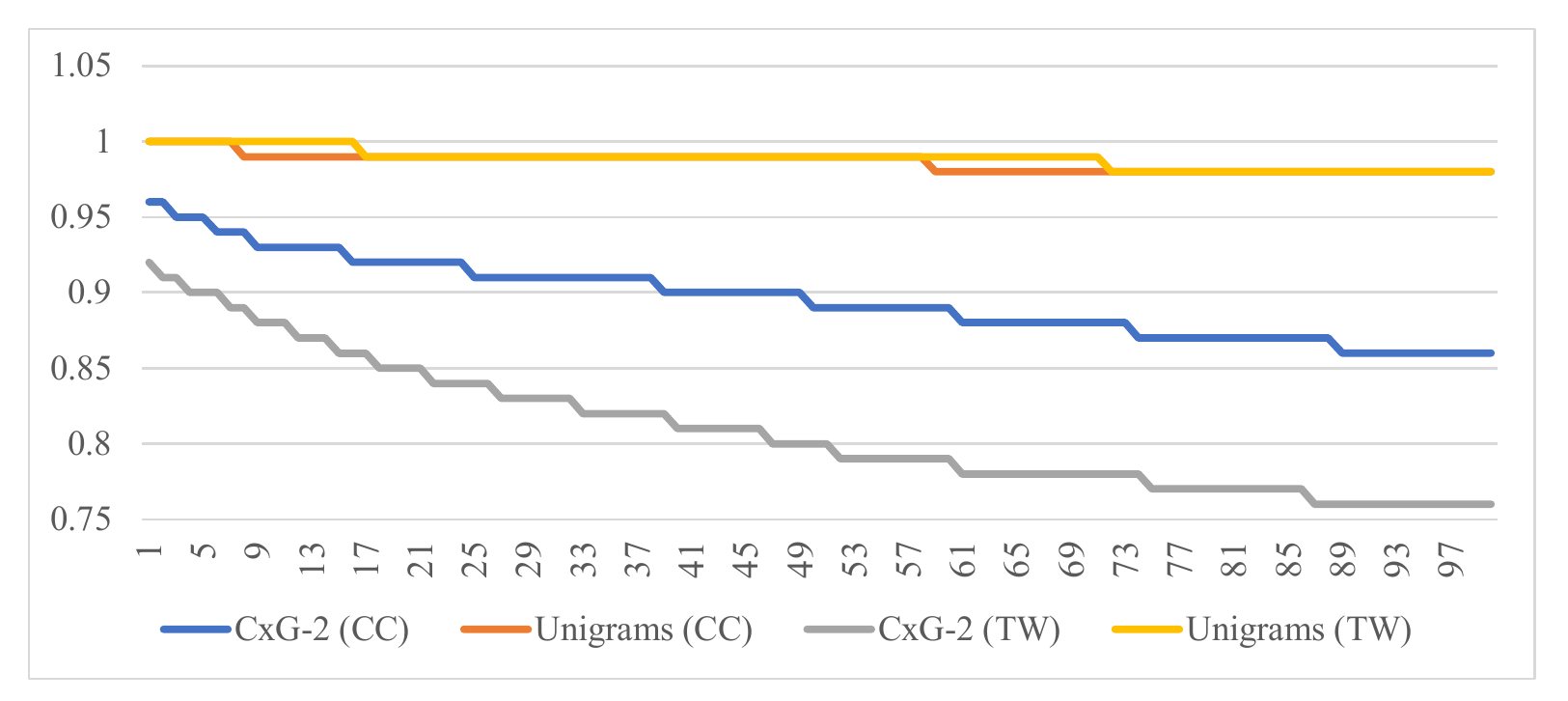}
\caption{Performance Over 100 Rounds of Unmasking (F1)}
\end{figure*}

This unmasking curve is simply a method for visualizing the robustness of a classification model. The syntactic model is less robust to unmasking than the lexical model. At the same time, we know that the syntactic model does not rely on place-names and place-based content and thus represents a more traditional linguistic approach to variation.

\section{Discussion}

This paper has used data-driven language mapping to select national dialects of English to be included in a global dialect identification model. The main experiments have focused on a dynamic syntactic feature set, showing that it is possible to predict dialect membership within-domain with only a small loss of performance against lexical models. This work raises two remaining problems:

First, we know that location-specific content (i.e., place names, place references, national events) can be used for geo-location and text-based models of \textit{place}. To what degree does a lexical approach capture linguistic variation (i.e., ``pop" vs. ``soda") and to what degree is it capturing non-linguistic information (i.e., ``Melbourne" vs. ``London")? This is an essential problem for dialect identification models. A purely syntactic model does not perform as well as a lexical model, but it does come with more guarantees.

Second, we have seen that inner-circle varieties have higher feature densities given the grammars used here. This implies that there are syntactic constructions in varieties like Philippines English that have not been modeled by the grammar induction component. While dialect identification can be used to optimize regional weights for \textit{known} constructions, how can such \textit{missing} constructions be adapted? This remains a challenge. While the less proto-typical dialects have higher F1s (i.e., Pakistan English), they also have lower feature densities. This indicates that some of their constructions are missing from the grammar. Nevertheless, this paper has shown that a broader syntactic feature space can be used to model the difference between many national varieties of English.

\nocite{*}
\bibliography{VarDial_19}
\bibliographystyle{acl_natbib}

\end{document}